# Sequential Classification of Aviation Safety Occurrences with Natural Language Processing

Aziida Nanyonga[1], Hassan Wasswa[2], Ugur Turhan[3], Oleksandra Molloy[4], Graham Wild.[5]
*University of New South Wales, Canberra, ACT, 2612, Australia*

Safety is a critical aspect of the air transport system given even slight operational anomalies can result in serious consequences. To reduce the chances of aviation safety occurrences, accidents and incidents are reported to establish the root cause, propose safety recommendations etc. However, analysis narratives of the pre-accident events are presented using human-understandable, raw, unstructured, text that a computer system cannot understand. The ability to classify and categorise safety occurrences from their textual narratives would help aviation industry stakeholders make informed safety-critical decisions. To classify and categorise safety occurrences, we applied natural language processing (NLP) and AI (Artificial Intelligence) models to process text narratives. The study aimed to answer the question, "How well can the damage level caused to the aircraft in a safety occurrence be inferred from the text narrative using natural language processing?" The classification performance of various deep learning models including LSTM, BLSTM, GRU, sRNN, and combinations of these models including LSTM+GRU, BLSTM+GRU, sRNN+LSTM, sRNN+BLSTM, sRNN+GRU, sRNN+BLSTM+GRU, and sRNN+LSTM+GRU was evaluated on a set of 27,000 safety occurrence reports from the NTSB. The results of this study indicate that all models investigated performed competitively well recording an accuracy of over 87.9% which is well above the random guess of 25% for a four-class classification problem. Also, the models recorded high precision, recall, and F1 scores above 80%, 88%, and 85%, respectively. sRNN slightly outperformed other single models in terms of recall (90%) and accuracy (90%) while LSTM reported slightly better performance in terms of precision (87%). Further, GRU+LSTM and sRNN+BLSTM+GRU recorded the best performance in terms of recall (90%), and accuracy (90%) for joint models. These results suggest that the damage level can be inferred from the raw text narratives using NLP and deep learning models.

## I. Nomenclature

*NLP*   = Natural Language Processing
*LSTM*  = Long Short-Term Memory
*NTSB*  = National Transportation Safety Board
*Seq2seq* = Sequence to sequence
*ASRS*  = Aviation Safety Reporting System
*ML*    = Machine Learning

---

[1] PhD student, School of Engineering, and Information Technology.
[2] PhD student, School of Engineering, and Information Technology.
[3] Senior Lecturer, School of Engineering, and Information Technology.
[4] Lecturer, School of Engineering, and Information Technology.
[5] Senior Lecturer, School of Engineering, and Information Technology.

*CNN*    =   Convolutional Neural Network
*RNN*    =   Recurrent Neural Networks
*sRNN*   =   Simple Recurrent Neural Network
*GRU*    =   Gated Recurrent Unit
*BLSTM*  =   bidirectional long short-term memory
*MLR*    =   multiple linear regression
*ANN*    =   artificial neural networks

## II. Introduction

Air transport is among the most sensitive fields to operational errors in the transport industry. It is a highly safety-critical industry where even the slightest mistake or minor system misconfiguration can result in a serious disaster whose consequences may include, though not limited to, loss of lives, huge financial losses, loss of customer trust, and property among others [1]. When an aviation safety incident occurs, be it minor, serious, or fatal, reporting and investigations are required. The main goal of this reporting on safety occurrences is not to assign blame but to ensure that similar events are prevented.

The report from aviation safety occurrence, either investigations or self-reports, containing unstructured text narratives, in a human-understandable language [2-5], highlighting the series of events that can be considered probable causes of the incident/accident. Also, these reports are freely available to the public. The goal of publishing these reports is partly to enable various stakeholders in the aviation industry ranging from the maintenance teams, operational manager to researchers in the aviation safety field access, analyze, critique, and implement/follow the recommended safety standard operating procedures, rules, and regulations.

However, in the event of a similar occurrence, investigation teams tend to revisit the reports from the previous occurrences. This can be extremely time-consuming if the conditions in which the incident in question occurred relate to conditions from many previous incidents. This is because, despite most reports including a tabular summary of the conditions in which the event occurred to make them more readable to humans, the actual detailed series of events that tend to be the probable incident/accident causes are usually given as a narrative using some natural language understood by the human stakeholders.

In addition, when reporting such narratives, humans often use non-standard terms that can be understood by fellow humans but not computers. This implies that, in case of a need to revisit these reports, humans will have to search through a large database of reports and sequentially read each of the related reports in a one-by-one approach. Consequently, this manual approach delays the investigation process. To mitigate this challenge, various models have been proposed to accelerate aviation data processing and aid in safety-critical decisions by the various aviation authorities and stakeholders. Approaches such as the famous multiple linear regression (MLR) have seen a wide deployment by numerous studies [6,7] to propose models that predict, and consequently plan for, the air transport market to maximize its financial and economic related benefits.

In addition, state-of-the-art models have been proposed to predict the same, using both traditional and modern/advanced machine learning algorithms such as Artificial Neural Networks (ANN) [2, 8] Random Forest (RF), [9] Support Vector Machine (SVM) [10, 11] and many more. However, most of these studies have focused on structured data, paying little or no attention to the unstructured text narrative. A few studies that have deployed natural language processing to aviation incident reports, have mostly used models like Latent Dirichlet Allocation (LDA) [12] to execute topic modelling tasks in the aviation industry [12, 13]. Though topic modeling can be used to perform classification of safety data report into several topics, its efficient falls with increasing topic overlap/ambiguity as a set of many words can be assigned to more than one topic/class making it highly unreliable. Although, deep learning models have gained increased popularity and are embraced across aviation field [14]. NLP for spam text email detection [15] and in bioinformatics for breast cancer detection, myocardial infarction survival prediction [9], and many more, it is not yet fully embraced by the aviation industry vis-a-vis big data analysis.

In this work, we apply NLP on aviation incident/accident reports for text analysis and advanced deep learning algorithms including Long Short-Term Memory (LSTM), Bidirectional-LSTM (BLSTM), Gated Recurrent Unit (GRU), and Simple Recurrent Neural Network (SimpleRNN) for classification [16, 17]. We compare the classification performance of each algorithm against the rest of the algorithms used in this study. The underlying research question is, given a text narrative describing the series of safety occurrence events, can we predict the resultant aircraft damage level? We based our study on the categories given in the NTSB investigation reports where for any aviation accident/incident the damage level to the aircraft is classified as either destroyed, substantial, minor or none. This can help the aircraft maintenance team to make informed decisions e.g., whether the aircraft can be put back to operation or not and if yes, how much repair does it need and consequently, anticipate the require budget, and other resource requirements and financial implications to the aviation company. However, this can have profound consequences if a wrong classification is made and can lead to unrecoverable losses or even forcing the company out of business. The rest of this work is organized as follows. Section III gives an account of the existing literature regarding aviation safety and machine learning. Section IV gives details of the proposed approach, and the implementation procedure followed to realize it. In section V we present the results of the study and give a detailed discussion and their implication to aviation safety research. Section VI presents the conclusion highlighting the direction of future work.

## III. Related Work

According to [18] machine learning enables computers to learn and make rational decisions based on experience, action, and reaction. Machine Learning (ML) has been successfully used in many fields of aviation, medicine, bioinformatics, biology, and many others. The most significant application of machine learning is data mining and different NLP and machine learning (ML) techniques have been applied to aviation safety reports. This part looks at different recent studies on text classification and text mining techniques that have been applied on various occasions.

Different studies applied different RNN architectures on sentence modelling where CNN and their results shows that these models are more suitable when applied sequential dataset such text mining, [19, 20, 21] they applied RNN technique for predicting weather related task and they reported that with their results they were able to regulate information recorded before the flight take off, [22] looked at different NLP techniques on aviation reports his main focus was to examine all the existing NLP that has previously been used on specifically civil aviation corpus though the performance of those techniques were not clearly reported. His study recommended the use of NLP, especially RNN, on text mining of time series data.

In his study, [23] proposed a deep learning approach for obtaining meaningful narratives from aviation safety reports. His study analyzed 186,000 ASRS (Aviation Safety Reporting System) reports using word2vec model for identifying a set of similar terms and find semantic terms within those reports. His proposed model showed that it can help in improving the set of terms used by reporting experts if used in aviation reports and hence reduce the uncertainty of existing safety reports thus making it easy for experts to understand. Though his study focused on only one NLP tool hence and hence a need for an improvement using different NLP approaches.

In their study, [24] used sequential deep learning for aviation safety prognosis on NTSB reports. They developed classification models using LSTM and word embedding and reported that these models were good and would help aviation experts in reviewing and analyzing all safety investigation reports. Their proposed method was done in two parts where they first transformed data extract on NTSB dataset into label and DL models were later developed for prediction regarding accident and other events.

In another study in predicting excess Events in Aircraft engines [25], applied an LSTM model to prediction task using data from 85 flights with a total of 79557 seconds (about 22 hours) of data. They aimed to examine suitable RNN using LSTM architecture neurons in predicting vibrations of aircraft engine. After model training and testing from aircraft engine vibration dataset, the proposed model was able to predict vibration values. LSTM architecture was compared with traditional RNN, and it showed that LSTM RNNs (recurrent neural networks) can correctly predict the vibration on flight dataset.

The authors in [26] evaluated the ability of different ML techniques to determine critical causes of accidents in air transport and its impact. To be precise, their study used decision tree, Naïve Bayes, and Sequential minimal optimization (SMO) to classify aviation accidents. Their study classified several airplane accidents and after the analysis, the results showed that the decision tree algorithm was better in predicting the cause of accident. It showed that loss of attention (human factor) as the most cause of accident in aviation and this would help air transport experts to minimize the loss of lives. The focus of their study is based on structured datasets which unstructured datasets also need to be considered based on different ML algorithms.

In a partnership study between CLLE-ERSS research lab and CFH safety company, [4] carried out a description of several NLP tools and text mining methods that can appropriately be used on aviation report unstructured data. The techniques and experiments conducted in their research only based on natural language processing tools and they based on [27] that suggested that these NLP techniques are more important when it comes to accident report analysis hence reported that NLP can be so helpful at any level of accident and incident report starting from reporting the incident to the final stage of analysis. Their study compared NLP on different kinds of dataset from different databases including ASRS, ECCAIRS (European Coordination Centre for Accident and Incident Reporting Systems), the Accident/ Incident Data Reporting (ADREP), DGAC and CFH / Safety Data. These are found in the public domain for use with safety reports. These studies did not report the exact NLP tools they used, and it is not clear if some tools could be used for better performance.

In their work, [16] produced and tested different RNN models on a dataset from NASA's Ames and Langley Research Centers. Their research was examining and predicting F-16 fighter jet data driven modeling and the fault sensor condition. They used different RNN architectures, and this include simpleRNN (sRNN), gated recurrent unit and long-short-term memory (LSTM) and a combination of these RNN architectures was later carried out where GRU-LSTM, GRU-sRNN, and LSTM-sRNN was combined to each other to determine the best performing layer. The study based on the performance of these models to predict the angel of attach of F-16 flight jet, faulty of angel of attach measurements and many others. The study showed that sRNN using trained estimated algorithm (Adam) achieved better performance in predicting as compared to all other algorithms used and sRNN combined with GRU performed better than all other networks that were used in their study.

Their study [17] used BLSTM, CNN and RNN to analyze aviation accident narratives and their main aim was to improve the text classification accuracy with a better understanding of accident records. Their study showed that these models can increase the accuracy on accident dataset and a recommendation on using more models on accident narrative dataset on was made. Although the proposed model performed better but its performance was less than 80% and if different parameters were used it would yield better performance of these models.

According to reference [28] proposed a 4-step approach to construct a network based on Bayesian probability distribution that would infer the causal relationships between a series of aviation accidents. Their approach utilized aviation data collected from 1982 to 2006 as reported in the National Transportation Safety Board (NTSB) aviation accident database. However, although their approach captured most of the causal relationships and recorded commendable performance, the dataset used for their study is incredibly old and does not capture the current trends and features embedded in today's aviation data. This is attributed to the fact there has been great advancement in aviation technology which translates into datasets with different distribution patterns

In another study, [29] proposed a data management scheme that deployed ontologies and conceptual models to enhance aviation safety data management. The main aim was to improve civil aviation management, repair, and overhaul maintenance. However, their approach focused on modelling conceptual frameworks and does not deploy any advanced data analytics or machine learning techniques. Though the proposed scheme recorded commendable performance, it does not adapt to the dynamic nature of the current aviation environment with big data being the denominator of most operations.

In their work, [30] used PCA (Principal Components Analysis), a flight incident prediction model based on deep brief networks and principal component analysis. The deep brief network architecture has the benefit that each layer learns a separate set of complex features than the layer before. PCA on the other hand, that has proved to be handy in dimensional reduction, it is blind to data classes and often leads to reduced classification performance. This would degrade aviation safety whose consequences may be extremely catastrophic.

## IV. Methodology

To realize the proposed approach, phases including training data set identification, text processing, and classification were executed as shown in Figure 1. The next subsection gives a brief description of the data sets used in this study.

### A. Data Acquisition

Various bodies, including ATSB (Australian Transport Safety Bureau), ASRS, and NTSB, among others collect and publish aviation incident/accident investigation reports. For this study, we used the NTSB aviation incident/accident investigation reports. Depending on the nature of the problem in question, this dataset can be accessed from NTSB website, together with the meta data in various forms including individual monthly published .pdf reports, .json files, or by querying individual reports online or a .csv summary. In this work we downloaded a .json file containing incident/accident investigation details for the years 2005 to 2020. Also, for the purpose of this study we considered incidents whose investigations were completed resulting in a dataset with 16919 records distributed among the four classes as 1409 destroyed, 15163 substantial, 195 minor and 152 with no damage to the aircraft. For each report we extracted the 'analysis Narrative' and 'damage Level' fields for training and validation of our deep learning models.

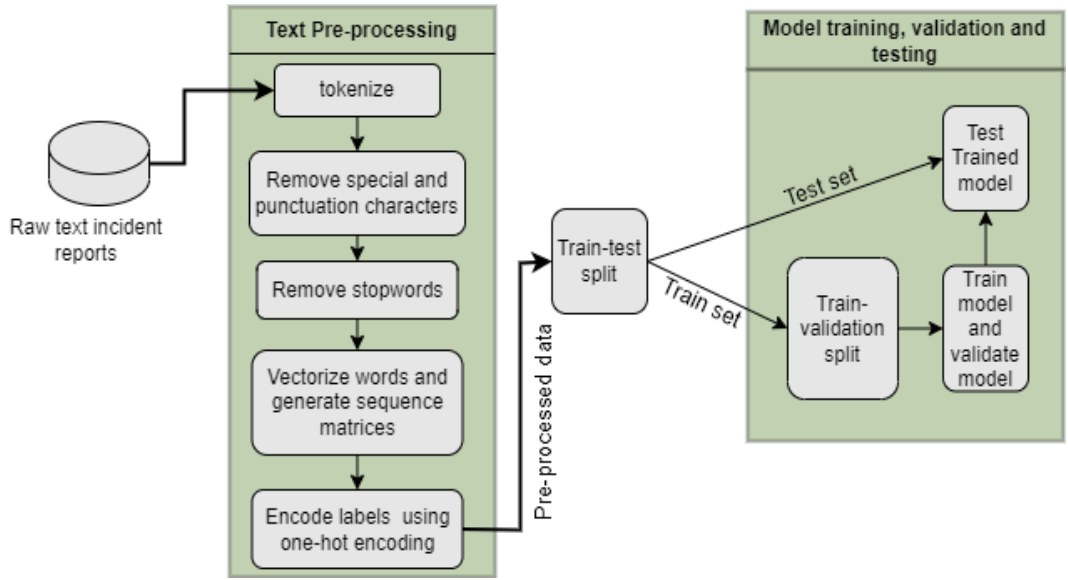

**Fig. 1 Methodological framework.**

### B. Text Processing

Machine leaning models are not designed to understand raw text data and therefore text data must be transformed from the human readable form to a numeric form that can be understood by the models. The Keras deep learning library was used since it hosts a whole bunch of deep learning models, and the various variants of model layers used in in this study. In addition, it provides some advanced modules for text pre-processing including the Tokenizer module which generates tokens and the sequence vectors for the input text, the categorical module which maps each of the unique categorical entries of the damage level variable (i.e., destroyed, Substantial, Minor and None) to a unique numerical entry or as a one-hot encoded entry for every data instance. For unwanted special and punctuation characters, and stop-words removal, and word lemmatization, the spacy library was used. Spacy is a Python library designed specifically to perform text processing tasks like named entity recognition, and word tagging and also incorporates a comprehensive list of special characters, punctuation and stop-words and is regularly updated whenever need arises. Using the above tools, each of the input text narrative was processed and transformed into a representative sequence/vector of length 2000. Numeric sequences generated from text narratives with number of words less than 2000 are padded with zeros while those whose length exceeds 2000 are truncated. The corpus vocabulary was set to 100,000. Scikit-learn's train-test-split module is deployed for splitting the data set into training, validation, and testing sets. All experiments are conducted using Python as the programing language and Jupyter notebook as a code editor.

### C. Text Classification

The processed data set is then randomly split as 80% train, and 20% test sets. In addition, 10% of the train set is held out for each epoch for model validation during training. Deep learning models including CNN, LSTM, BLSTM, GRU, SimpleRNN, and combinations of these models including LSTM+GRU, BLSTM+GRU, SimpleRNN+LSTM, SimpleRNN+BLSTM, and SimpleRNN+GRU are trained, and their performance evaluated and compared against other models. For model optimization, the Adam optimizer was deployed. However, this study was not concerned about finding the best optimizer and therefore any other optimization can be used.

### D. Deep learning model Architecture

To maintain uniformity, with the exception of slight variations for combined models, all models used a common architecture constituting an embedding layer, hidden layers and output layer. For all hidden layers, Rectified Linear Unit (ReLU) was deployed for activation while the output layer deployed SoftMax as the activation function. To generate the predicted class, the argmax function was used. This returns the index corresponding to the entry with highest probability from the SoftMax output. Figure2 shows the general deep learning architectures used in this study for a single RNN-based deep learning algorithm (a) and when two different RNN-based deep learning algorithms are combined (b).

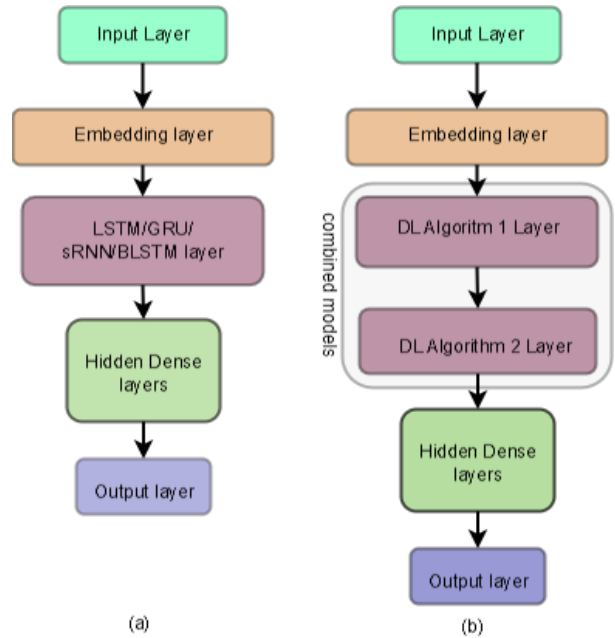

**Fig. 2 Deep learning architectures: (a) using single deep learning algorithm and (b) using two different Deep learning algorithms.**

## V. Results and Discussion

The results from the experiments conducted in this study showed that, using natural language processing together with advanced/modern deep learning models, one can, with a high degree of accuracy and competitive precision, predict degree of impact to the aircraft in terms of damage level given the unstructured text narrative of the pre-accident series of events. In this section we present and discuss the experimental results obtained for each deep learning model used accompanied with a precise description of their implications. As already indicated in section, we compared the performance of different deep models for classification of accident narrative reports. The architectures were evaluated in two perspectives, i.e., single model architecture where a single recurrent neural network layer was used for instances GRU, sRNN, LSTM, and BLSTM and a joint RNN architecture where the model constituted layers of two or more RNN models for instance GRU with LSTM, BLSTM, sRNN as GRU-LSTM, GRU-LSTM, GRU-sRNN and sRNN with LSTM, BLSTM as BLSM-sRNN, LSTM-sRNN.

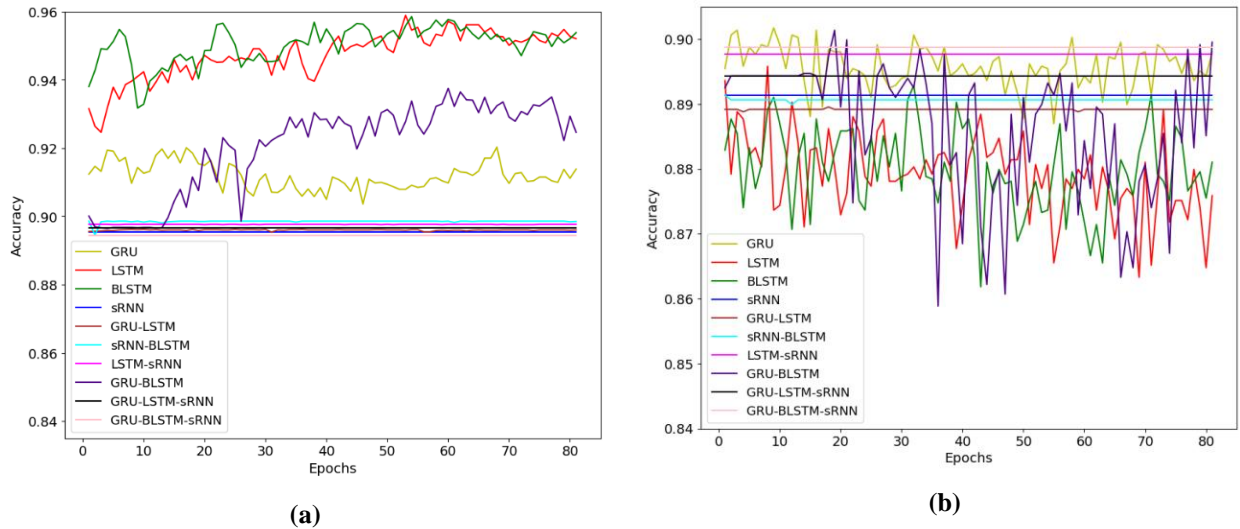

**Fig. 3 Single RNN-based Model performance in terms of training (a) and validation (b) accuracy.**

## A. Detection performance of single RNN-based deep learning models

In this section, the individual performance of each model is presented and compared with others in terms of training accuracy, validation accuracy, testing accuracy, precision, recall and F1 measure. It is worth noting that in the context of this study the term single RNN-based deep learning model refers to a deep learning model that embeds only one RNN-based layer (see Figure 2.( a)) while joint RNN-based deep learning model refers to a deep learning model that embeds two or more RNN-based layers with the output of one layer serving as input for the subsequent layer of a different RNN architecture (see Figure 2.(b)). Table 1 shows the performance of each of the four single RNN-based deep learning models used in this study. Apart from testing performance the training and validation results are analyzed and presented in Figure 3.

**Table 1. Single RNN-based Deep learning model performance**

| Modals | Precision (%) | Recall (%) | F1-Score (%) | Accuracy (%) |
|---|---|---|---|---|
| LSTM | **87** | 89 | **88** | 88.9 |
| BLSTM | 85 | 88 | 86 | 87.9 |
| sRNN | 82 | **90** | 86 | **90.0** |
| GRU | 85 | 89 | **88** | 89.0 |

It can be seen from Table 1 that all models perform competitively well with simple recurrent neural network (sRNN) slightly outperforming others in terms of recall and accuracy while LSTM reported slightly better performance in terms of precision. This result, indicates that, the damage level caused to the aircraft can, with an accuracy far above random guess, be predicted from the narrative of pre-accident series of events given that the unstructured narrative is appropriately processed and transformed into a correct state for machine learning models.

## B. Detection performance of joint RNN-based deep learning models

Aside from single RNN-based deep learning models, joint RNN-based models were evaluated on the same data set. Table. 2 presents the prediction results from the six different combinations of the four RNN-based models evaluated in this study. The training and validation results of joint models are visualized in Figure 4. A comparison of the test performances between single models and joint models in terms of accuracy, precision, recall and F1-score is shown in Fig 5.

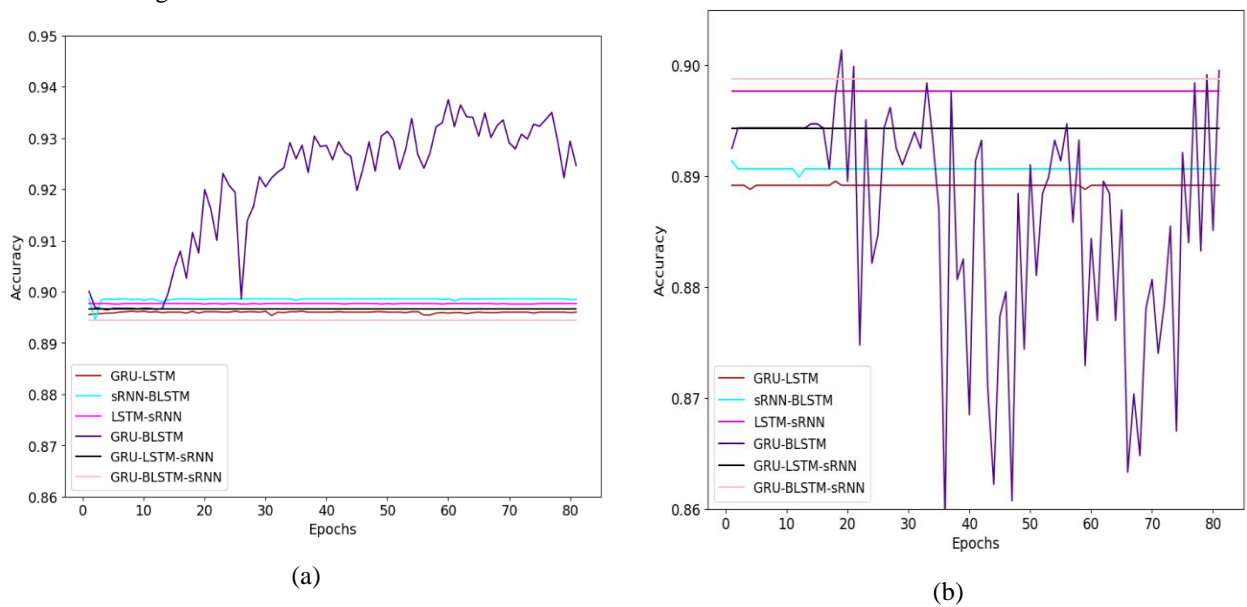

**Fig. 4 Joint RNN-based Model performance in terms of training (a) and validation (b) accuracy.**

**Table 2.** Joint RNN-based Deep learning model performance

| Modals | Precision (%) | Recall (%) | F1-Score (%) | Accuracy (%) |
| --- | --- | --- | --- | --- |
| GRU- LSTM | 81 | 90 | 86 | **90.0** |
| GRU- BLSTM | **88** | 88 | 88 | 88.1 |
| sRNN-BLSTM | 81 | 90 | 85 | 89.8 |
| sRNN-LSTM | **88** | 89 | **89** | 89.0 |
| GRU-BLSTM- sRNN | 81 | 90 | 85 | **90.0** |
| GRU-LSTM-sRNN | 80 | 90 | 85 | 89.6 |

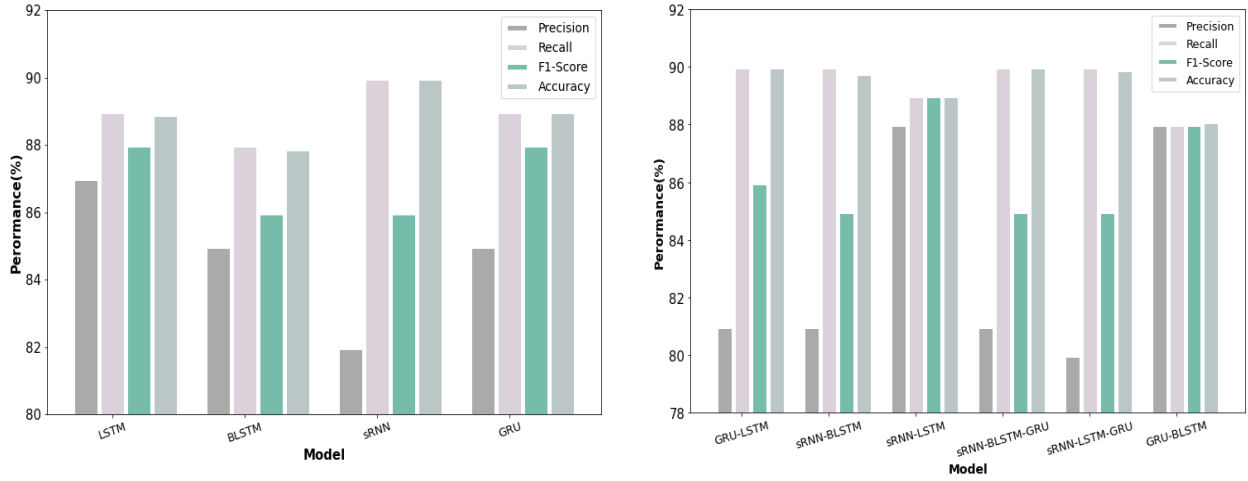

**Fig. 5 Test Performance Visualization of all modals.**

## C. Single RNN-based versus Joint RNN-based deep learning models

In this section we present the analysis of the two classes of models studied in this work. Table 3 presents the performance for all the models evaluated in this study. Although the individual performance for each of the studied models is satisfactorily competitive, generally, the joint RNN-based models outperformed the single RNN-based models with, considering general performance across metrics, sRNN-LSTM showing the overall best performance.

**Table 3.** Model performance

| Methods | Precision (%) | Recall (%) | F1-Score (%) | Accuracy (%) |
| --- | --- | --- | --- | --- |
| LSTM | 87 | 89 | 88 | 88.9 |
| BLSTM | 85 | 88 | 86 | 87.9 |
| sRNN | 82 | **90** | 86 | **90.0** |
| GRU | 85 | 89 | 88 | 89.0 |
| GRU- LSTM | 81 | **90** | 86 | **90.0** |
| GRU- BLSTM | **88** | 88 | 88 | 88.1 |
| sRNN-BLSTM | 81 | **90** | 85 | 89.8 |
| sRNN-LSTM | **88** | 89 | **89** | 89.0 |
| GRU-BLSTM- sRNN | 81 | **90** | 85 | **90.0** |
| GRU-LSTM-sRNN | 80 | **90** | 85 | 89.9 |

## VI. Conclusion

Safety is one of the core components in the aviation industry and therefore requires dire monitoring accompanied with critical observation of the standard operating procedure, safety rules, regulation, and recommendations from the field experts. Often the reports from investigations about aviation anomalies that threaten the safety of stakeholders at all levels, are given with a textural narrative describing the series of events. In this paper we leveraged various NLP techniques and various deep learning models to classify the damage level of the aviation accident to the aircraft based on textual accident/incident narratives from the NTSB investigation reports. With considerably competitive results from our experiments, we answered the question of whether the damage level can be inferred from the raw text narratives.